\documentclass[conference]{IEEEtran}
\IEEEoverridecommandlockouts

\usepackage{cite}
\usepackage{amsmath,amssymb,amsfonts,amsbsy}
\usepackage{algorithmic}
\usepackage{graphicx}
\usepackage{textcomp}
\usepackage{xcolor}
\usepackage{gensymb}
\usepackage{hyperref}
\usepackage{mwe}
\usepackage{verbatim}
\usepackage{siunitx}
\usepackage{tabularx, booktabs}
\usepackage{microtype}

\def\BibTeX{{\rm B\kern-.05em{\sc i\kern-.025em b}\kern-.08em
    T\kern-.1667em\lower.7ex\hbox{E}\kern-.125emX}}

\begin{document}

\title{Deep Reinforcement Learning Using a\\ Low-Dimensional Observation Filter for \\ Visual Complex Video Game Playing}

\makeatletter
\newcommand{\linebreakand}{%
  \end{@IEEEauthorhalign}
  \hfill\mbox{}\par
  \mbox{}\hfill\begin{@IEEEauthorhalign}
}
\makeatother

\author{\IEEEauthorblockN{Victor Augusto Kich}
\IEEEauthorblockA{\textit{Universidade Federal de Santa Maria} \\
Santa Maria, Brazil \\
victorkich@yahoo.com.br}
\and
\IEEEauthorblockN{Junior Costa de Jesus}
\IEEEauthorblockA{\textit{Universidade Federal de Rio Grande} \\
Rio Grande, Brazil \\
dranaju@gmail.com}
\and
\IEEEauthorblockN{Ricardo Bedin Grando}
\IEEEauthorblockA{\textit{Universidad Tecnológica del Uruguay} \\
Rivera, Uruguay \\
ricardogrando13@gmail.com}
\linebreakand % <------------- \and with a line-break
\IEEEauthorblockN{Alisson Henrique Kolling}
\IEEEauthorblockA{\textit{Universidade Federal de Santa Maria} \\
Santa Maria, Brazil \\
alikolling@gmail.com}
\and
\IEEEauthorblockN{Gabriel Vinícius Heisler}
\IEEEauthorblockA{\textit{Universidade Federal de Santa Maria} \\
Santa Maria, Brazil \\
gvheisler@inf.ufsm.br}
\and
\IEEEauthorblockN{Rodrigo da Silva Guerra}
\IEEEauthorblockA{\textit{Universidade Federal de Santa Maria} \\
Santa Maria, Brazil \\
rodrigo.guerra@ufsm.br}
}

\maketitle

\begin{abstract}

Deep Reinforcement Learning (DRL) has produced great achievements since it was proposed, including the possibility of processing raw vision input data. However, training an agent to perform tasks based on image feedback remains a challenge. It requires the processing of large amounts of data from high-dimensional observation spaces, frame by frame, and the agent's actions are computed according to deep neural network policies, end-to-end. Image pre-processing is an effective way of reducing these high dimensional spaces, eliminating unnecessary information present in the scene, supporting the extraction of features and their representations in the agent's neural network. Modern video-games are examples of this type of challenge for DRL algorithms because of their visual complexity. In this paper, we propose a low-dimensional observation filter that allows a deep Q-network agent to successfully play in a visually complex and modern video-game, called Neon Drive. \footnote{Video available at: \texttt{\scalebox{0.85}{\url{https://youtu.be/bKdMBKu68QU}}}}\footnote{Code available at: \texttt{\scalebox{0.85}{\url{https://github.com/victorkich/Neon-Drive-DRL}}}}

\end{abstract}

\section{Introduction}

Although earlier games used much simpler algorithms, and the definition of AI shifts according to the era, it could be said that video games have been incorporating AI since the first Atari games -- see~\cite{skinner2019icccs} for a review. AI has also aided in the improvement of the way humans play, understand, and build games \cite{yannakakis2018artificial}. More recently, many studies began investigating how an artificial intelligence that is external to the game itself, can be used to play it at a human level or beyond, while being subjected to the same boundaries in terms of perception feedback and controls. Video games are ideal contexts for AI research benchmark because they present intriguing and complicated problems for agents to solve, and these problems are defined in controlled and repeatable environments that are secure and easy to manage. 
\begin{figure}[ht]
\centering
\includegraphics[width=\linewidth]{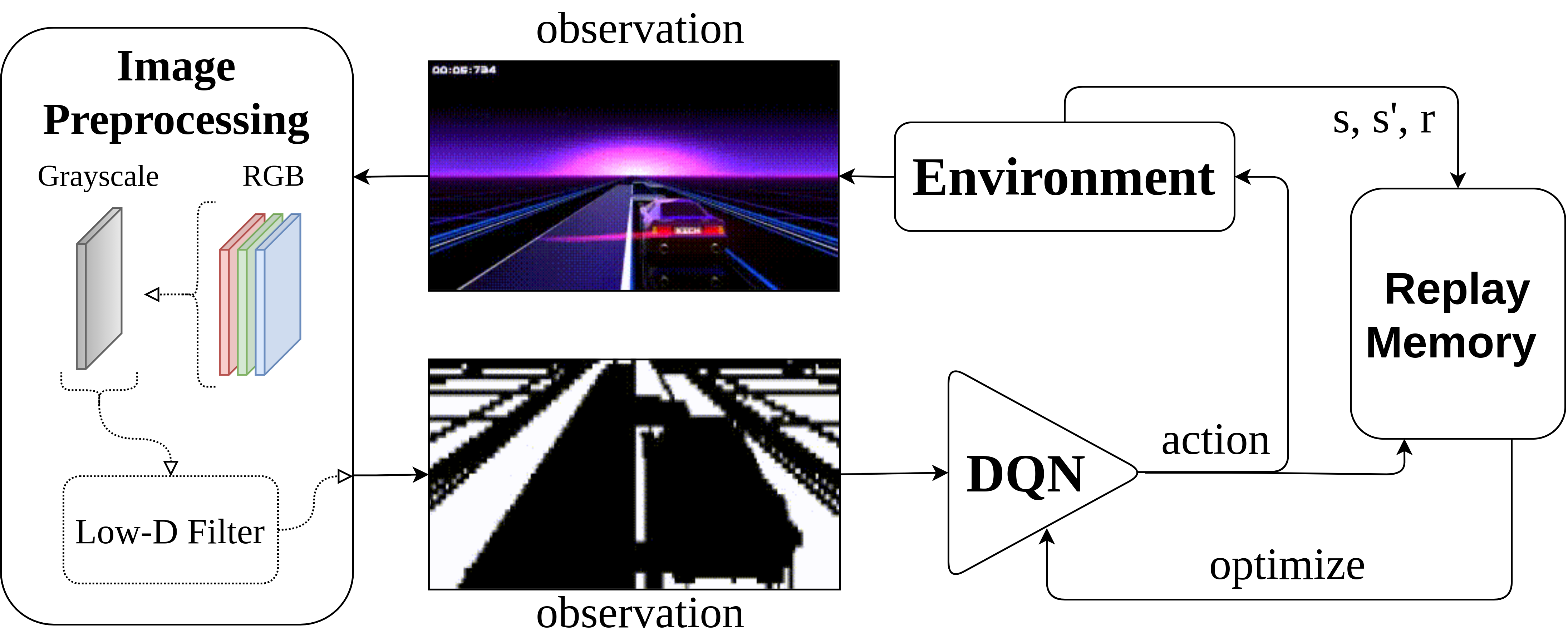}
\caption{System structure used to train the DQN agent using a low-dimensional image as observation for the input.}
\label{fig:system_diagram}
\vspace{-6mm}
\end{figure}
Furthermore, games provide an abundant source of usable data for Machine Learning (ML) algorithms and often run significantly faster than real-time. Games are a unique and popular topic for AI study because of these properties.

Since Chess and Go have been effectively conquered, the awareness that video games are good benchmarks for artificial intelligence methods has been established throughout the AI field. Since DeepMind's landmark paper showing that Q-learning combined with Convolutional Neural Networks (CNN) could learn to play many of the Arcade Learning Environment (ALE) games at a superhuman level \cite{mnih2015human} there has been an almost daily flurry of new papers applying AI approaches to video games. The ALE, which is based on an emulator for the Atari 2600 games console and contains dozens of games \cite{bellemare2013arcade}, has been used in numerous published papers.

In general, game AI is concerned with sensing and decision-making in virtual worlds. There are some critical difficulties and possible solutions associated with these components. Here we highlight two of such difficulties. The first is that the game's state space is usually very wide. Such large-scale state-spaces were successfully modelled with Deep Neural Networks (DNN) thanks to the rise of representation learning. The second, is that it is difficult to learn correct policies for making decisions in a dynamic, uncertain environment. Data-driven methods, such as supervised learning and Deep Reinforcement Learning (DRL), are viable solutions for this problem.

In this work, we demonstrate the effectiveness of DRL on a visually complex modern video game, focusing mainly on the low-dimensional observation filter for preprocessing the input image of each time step. Our proposed approach consists of training our Deep Q-Network (DQN) \cite{mnih2015human} agent using our image filter. Our system definition can be seen in Fig.~\ref{fig:system_diagram}.

\section{Related Works}

The current literature of DRL applied in classic video games is wide, approaching a large set of problems. Super-human skill levels have already been achieved in several classic video games using DRL, demonstrating how effectively the approach can be.

Arulkumaran et al.~\cite{arulkumaran2017deep} provided a survey discussing the use of raw pixels from images to train DRL agents, correlating the historical progress of DRL in both video games and robotics, using visual inputs for agents to better understand their surroundings, enabling them to learn even high-level causal relationships. Utilizing pixels observations together with data augmentation, Laskin et al.~\cite{laskin2020reinforcement} manage to obtain results on par and better than DRL utilizing states observation, showing the benefits of such approach. 

Torrado et al.~\cite{torrado2018deep} trained some types of DRL algorithms, including the DQN, in 2D classical environments used in The General Video Game AI Competition (GVG-AI). The authors trained different agents in 160 different environments using only pixels as input, demonstrating the actual difficulty faced by the agents when trying to correlate one game to another.

Ha and Schmidhuber~\cite{ha2018world} used a variational autoencoder to reduce the dimensions of the observation space using the latent space provided by the encoder as input for training the controller in a 2D gym environment. In another work~\cite{ha2018recurrent} the same authors further developed the idea, training the agent in an imaginary environment, inside of its simulated latent space world. This approach spent much less time training to solve the same 2D gym environment.

Shao et al.~\cite{shao2019survey} promoted a more current review about DRL using video games as training environments, including Starcraft~2~\cite{vinyals2017starcraft}, Dota~2~\cite{berner2019dota}, Minecraft~\cite{tessler2017deep}, and others. This work also talks about the influence of optimization in the training process, highlighting how much hardware intensive is the training of such complex agents.

Lin et al.~\cite{lin2019using} replace the conventional CNN in favor of a deep principal component analysis network in order to reduce the dimensions of the observation space. Lin also uses a DQN in both Flappy Bird and Atari Breakout games, achieving a super-human performance. Tan et al.~\cite{tan2019modeling} use an image preprocessing approach to reduce the dimensional observation space, using image resizing followed by grayscaling and binarization, applied to a part of the image. Tan uses a DQN method to train the agent in the race car, a classic 2D gym environment. 

\section{Theoretical Background}
\label{theoretical}

It is difficult, in general, to train an agent to make decisions given high-dimensional inputs. It was empirically observed that RL from high-dimensional (state) observations with raw pixel images are sample inefficient \cite{kaiser2019model}. With the introduction of Deep Learning (DL), researchers employ DNN as function approximators to optimize policies.

\subsection{Deep Learning}

DL is a data representation learning methodology based on artificial neural networks (NN). It is based on the concept of brain development and can be taught in three different ways: supervised, unsupervised, or semi-supervised. Although the term ``deep learning'' was developed in 1986 \cite{dechter1986learning}, some argue the field is still in its infancy due to a lack of sufficient amount of data and capable enough computer hardware. However, as larger-scale datasets and more capable hardware become available, DL undergoes significant transformations \cite{schmidhuber2015deep}.

CNN \cite{krizhevsky2012imagenet} is a type of DNN frequently used in computer vision. CNN is built on shared-weights architecture and is shift-invariant, as it is inspired by biological processes. Speech recognition, image classification and segmentation, semantic comprehension, and machine translation are just a few of the disciplines where CNN systems have seen substantial success \cite{lecun2015deep}. %Traditional ML approaches can be surpassed by DL-based methods that use efficient parallel distributed computing resources. This strategy motivates scientists and researchers to strive for ever-higher levels of excellence in their areas.

\subsection{Reinforcement Learning}

\begin{figure*}[t]
\centering
\includegraphics[width=0.95\linewidth]{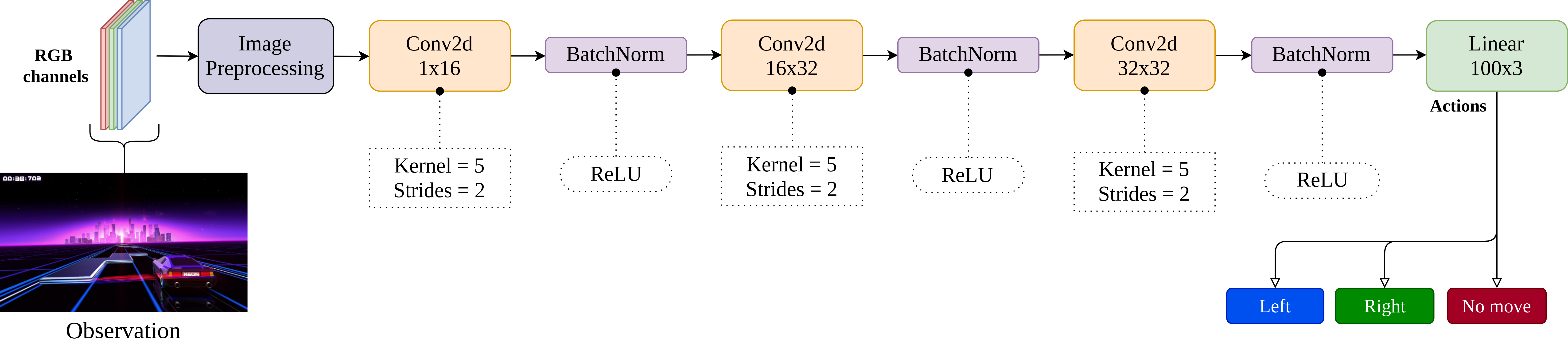}
\caption{Used structure to train the DQN agent with the low-dimensional observation filter. In our work, the observations generated by the environment can have any resolution because of the resize label before the CNN.}
\label{fig:structure_diagram}
\vspace{-4mm}
\end{figure*}

Reinforcement Learning (RL) is a type of ML method based on the idea of operand conditioning, in which agents learn the best policy through experimentation, exploring a state space by choosing actions and observing outcomes in order to model and maximize future rewards \cite{sutton2018reinforcement}. RL may be successfully applied to sequential decision-making tasks by interacting with the environment. Considering a discounted episodic Markov Decision Process (MDP) $(S,A,\gamma, P,r)$, the agent choose an action $a_t$ in the state $s_t$ based on the policy $\pi(a_t|s_t)$. The action is received by the environment, which generates a reward $r_{t+1}$ and transfers the agent to the next state $s_{t+1}$ $(s')$ based on the transition probability $P(s_{t+1}|s_t, a_t)$. The operation is repeated until the agent reaches a terminal condition or exceeds the maximum time step. The purpose is to maximize the projected discounted cumulative rewards
\begin{equation}
    \mathbb{E}_\pi[R_t]=\mathbb{E}_\pi\left[\sum_{i=0}^\infty\gamma^i r_{t+1}\right].
\end{equation}
where $\gamma \in (0, 1]$ is the discount factor.

Off-policy and on-policy methods of RL can be distinguished. Off-policy RL algorithms are those in which the behavior policy used to select actions differs from the one employed while learning the policy. On the contrary, in on-policy RL algorithms, the behavior policy is the same as the one used while learning. Furthermore, RL can be separated into value-based and policy-based techniques. Agents in value-based RL update the value function to learn appropriate policies, whereas policy-based RL agents learn the policy directly.

Q-learning is an example of an off-policy value-based method. The Q-learning update rule is
\begin{equation}
    \delta_t=r_{t+1}+\gamma\text{arg}\ \underset{a}{\text{max}}Q(s_{t+1},a)-Q(s_t, a_t),
\end{equation}
\begin{equation}
    Q(s_t,a_t)\gets Q(s_t,a_t)+\alpha\delta_t.
\end{equation}
where $\delta_t$ is the temporal difference error, and $\alpha$ is the learning rate.

\subsection{Deep Reinforcement Learning}

DRL is the fusion of DL and RL, and has shown tremendous development since its inception. This section will introduce the DQN technique \cite{mnih2013playing}, a pure value-based DRL technique.

The classic DQN algorithm \cite{mnih2015human} succeeded to exceed humans in numerous games by establishing the agent's behavior. This system was able to estimate the agent's action on a human level utilizing only a raw pixel image as input. DQN uses the experience replay method to break the sample correlation and stabilize the learning process with a target Q-network.
A target network generates targets for the temporal-difference error that can regulate the learning and, as in the experience replay, improve even more the stability of the method.
The target network $(\theta^{-})$ contains a copy of the Q-network $(\theta)$, however, with ``soft'' updates: $\theta^{-} \leftarrow \tau\theta + (1-\tau)\theta^{-}$. The loss function of $\theta$ at iteration $i$ is
\begin{equation}
    L_i(\theta_i)=E_{(s,a,r,s')\sim U(D)}\left[\left(y^{DQN}_i-Q(s,a;\theta_i)\right)^2\right],
\end{equation}
with
\begin{equation}
    y^{DQN}_i=r+\gamma\underset{a'}{\text{max}}Q(s',a';\theta^{-}_i).
\end{equation}
DQN bridges the gap between high-dimensional visual inputs and actions.

The training is performed with $\epsilon$-greedy exploration (the agent chooses the best action with probability epsilon and chooses a random action otherwise), and no regularization is used. In particular, dropout is usually advised against in such regression settings.

It is noteworthy that the DQN method has already shown difficulty in learning using visually complex environments. Another problem with DQN is the fact that this method could only handle discrete action spaces, making it necessary to use other approaches to perform it in a continuous space \cite{lillicrap2015continuous}.

\section{Methodology}
\label{methodology}

In this section, we discuss our DRL approach and we detail the network structure of our agents. We also present the image preprocessing pipeline and the proposed reward function for the task the agent must accomplish autonomously.

\subsection{Proposed Approach}

Our environment is a visually modern game called Neon Drive, which has a great variety of colors and changes in luminosity and a discrete space in the action domain. This game has a very simple purpose, to deviate from fixed obstacles over time using three types of discrete actions: left, right, and no move. Because of that, the DQN method, based on recent articles, is a good option for solving our problem, training the discrete agent using only the pixels of the image as input.

The neural network architecture used to train and other significant steps of our DQN agent structure can be seen in Fig.~\ref{fig:structure_diagram}. We used a classical CNN architecture, with three layers of convolution and layers of batch normalization between them. The activation function was applied to the batch normalization output and the outputs from the linear layer are the three discrete actions that will be sent to be performed in the environment.

\subsection{Image Preprocessing}

The core idea of reducing the extra image information present in the environment observation revolves around cropping and binary filtering. All of the realized processes in this subsection have been made using python modules, in special: OpenCV for image filters, Torchvision for image transformations, and mss for screen capturing.

\begin{figure}[ht]
\centering
\includegraphics[width=\linewidth]{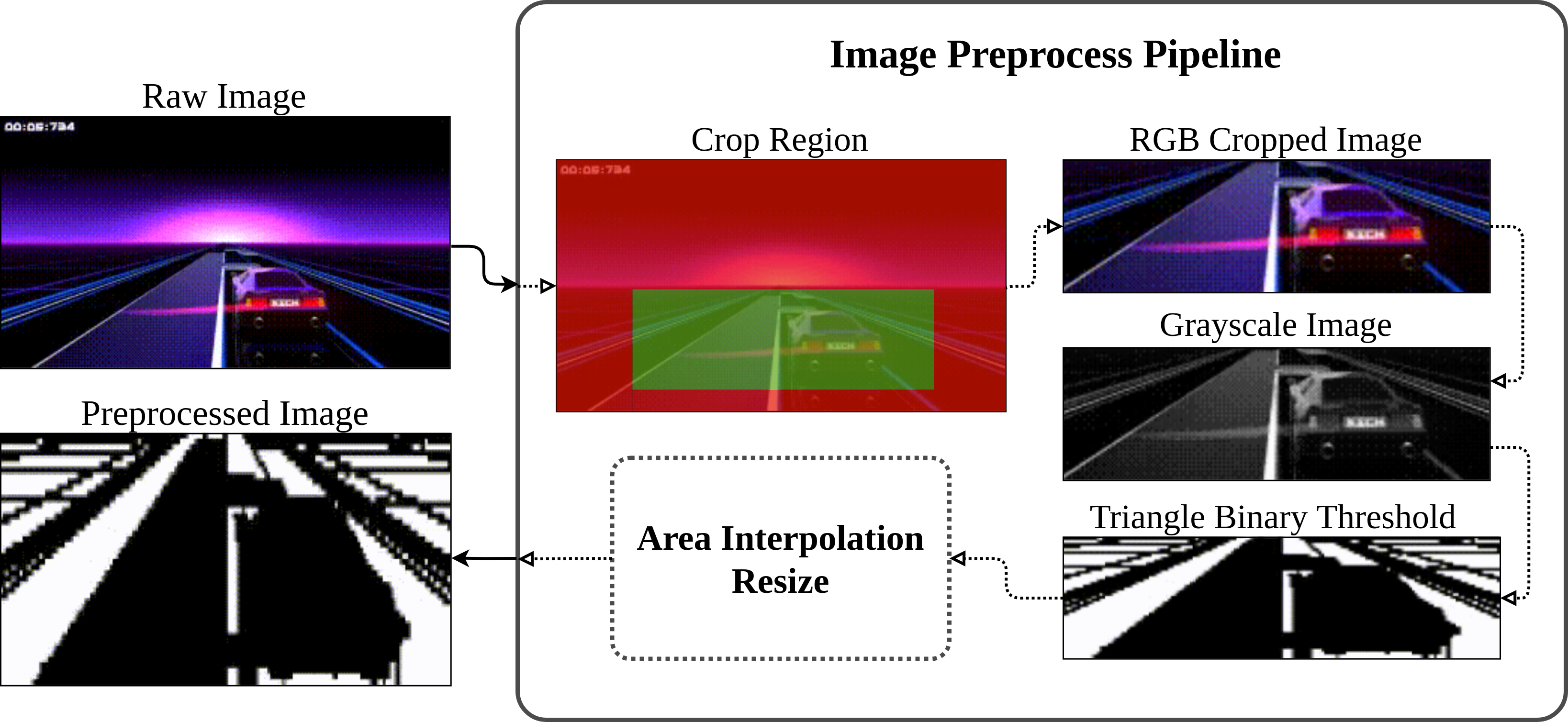}
\caption{The complete image preprocess pipeline was used to reduce the environment observation. Cropping the borders and removing the colors to binary black and white patterns.}
\label{fig:preprocess_diagram}
\vspace{-4mm}
\end{figure}

The image processing pipeline performed in this work can be seen in Fig.~\ref{fig:preprocess_diagram}. Through the mss python module, the screen was captured and transformed into an n-dimensional array variable. With the BGR screen saved, a color-to-grayscale filter was applied. Then the image was cropped, removing $53.84\%$ of the upper pixels, $20\%$ of the lower pixels, and $20\%$ of the left and right pixels. After that, the triangle threshold function was applied to transform the image to binary. Finally, we resize the final image to $160\times90$ pixels using area interpolation.

\subsection{Reward Function}

A reward and penalty functions must be established for the DRL strategy. Rewards and penalties are values assigned to the performance of the agent, allowing it to learn hyper-parameters through the feed-forward and back-propagation phases of the networks. A simple reward and penalty strategy was used, described as follows:

\begin{equation}
r(s_t, a_t)= 
\begin{cases}
    r_{alive}           & \text{if } d_o < d_t \\
    r_{collide}         & \text{if } [H(i_t)\sim H(i_f)]\leq0.15
\end{cases}
\end{equation}

If the car at the time $d_t$ has already surpassed the first obstacle in the $d_o$ a positive $r_{alive}$ reward of $1$ is given. If the car collides with any obstacle, a neutral $r_{collide}$ reward of $0$ is granted to the step, finishing the episode at this time. The collision is verified by the Bhattacharyya distance of two histograms $H$: if the similarity between the current raw image $i_t$ and the game over image $i_f$ is more than $85\%$, a collision has occurred.

\section{Experimental results}

The DRL approach was implemented using the Python programming language. The agent-related control modules were also implemented in Python. The implementation of the neural networks was carried out with the PyTorch\footnote{\texttt{\scalebox{0.85}{\url{https://pytorch.org/}}}} library. Both development and training were conducted on a computer with Ubuntu 20.04. The performance of our approach can also be observed in a complementary video.

\subsection{Training setup}

We used the proposed approach to train the agent for the Neon Drive in endurance mode. For all the training episodes, the agent's initial position was set at the start checkpoint. The episode ends if the vehicle collides with any of the obstacles present in the environment. This also means that the episode does not end if the agent does not hit an obstacle, so the total amount of reward of a given episode can reach a large value.

The neural networks have been trained with an RMSprop optimizer and with a learning rate of $10^{-3}$. The selected minibatch size was $128$. We set a limit of training of $4000$ episodes for the scenario, and the replay memory has size $30000$. The limit was defined empirically since good results could be observed around these values. 

\subsection{Results}

The use of $\epsilon$-greedy noise create a noisy reward graph, so the moving average of the reward is. depicted in Fig.~\ref{fig:reward}. It can be seen that it took approximately 1000 episodes to start noticing the agent was learning the task, yielding a considerably higher amount of reward by the end of the training.

\begin{figure}[h]
\centering
\includegraphics[width=\linewidth]{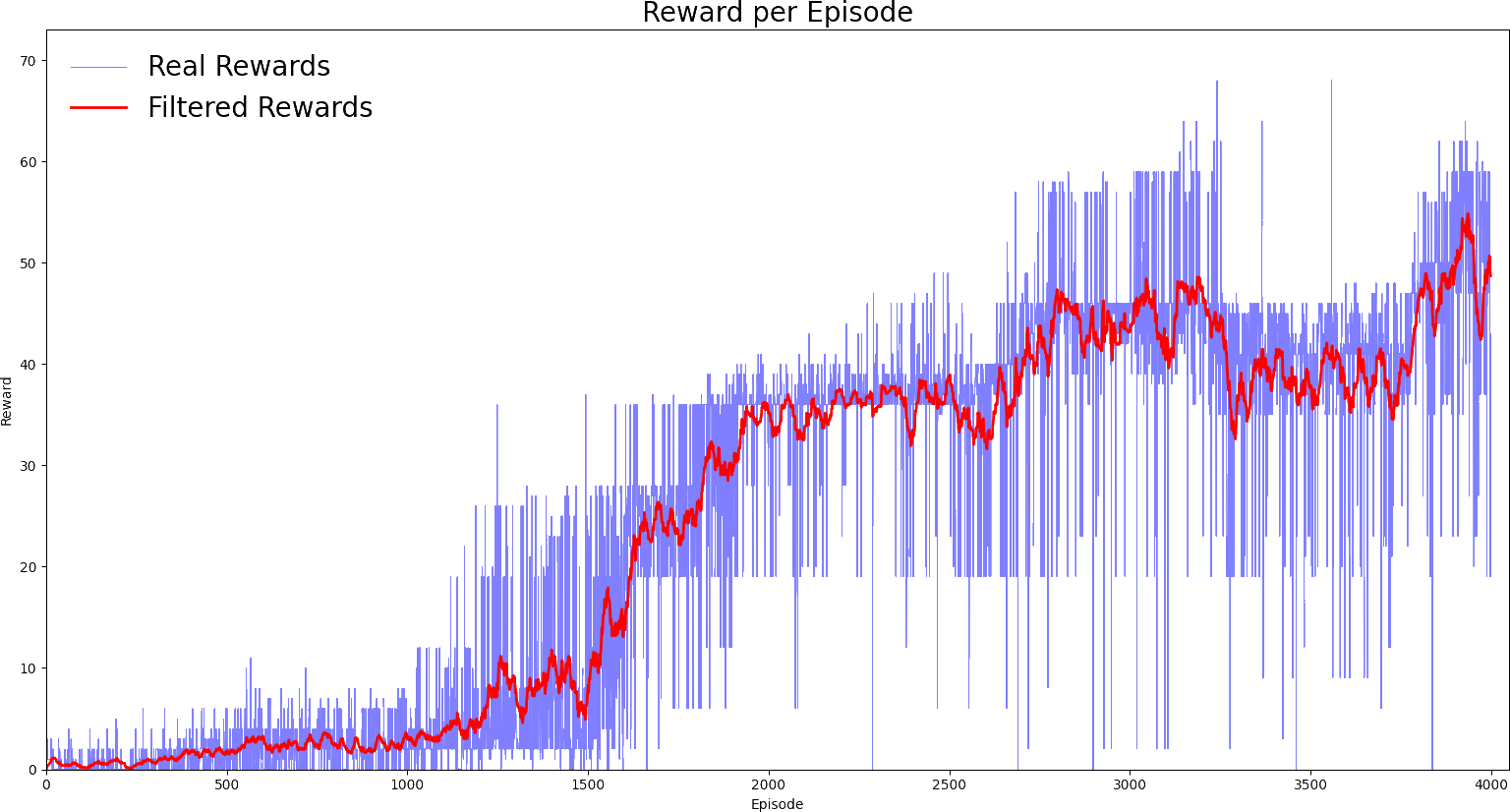}
\caption{Moving average of the reward of 4000 training episodes.}
\label{fig:reward}
\end{figure}

In our environment, the vehicle was set to always start at the same initial checkpoint. The agent had also to navigate through the environment, while avoiding the obstacles in order to maximize the reward, reaching the ``next goal" without a collision. Since the map never changed, the navigation was basically a memorization problem. The performance was measured by the moving average of 25 training episodes. This is shown in Fig.~\ref{fig:reward}.

Since our environment is a real game, the frame generation is uncontrollable, making the observation space even more stochastic. In our empirical tests, we defined our control frequency to generate 3 inputs per second. In order to take into consideration the temporal difference through the observation steps, we actually subtract the former observation from the current one, feeding the difference as the input -- except for the game over screen.

To test the robustness of our approach, we input salt \& pepper (S\&P) noise in the input images before the observation preprocessing \cite{bar2005image}. For this task, we use four samples of S\&P noise: $0.4\%$, $1\%$, $10\%$, and $25\%$. The results can be seen in Fig.~\ref{fig:reward_comparition}. For low noise inputs, the results kept almost the same average, but for the higher noise inputs, $10\%$ and $25\%$, the moving average of the reward is decreased. The delay between observations varies. The deliberate delay is $333ms$, however as the game continues to run, other system delays are added to the total, and also the initial loading time of the game varies a lot, all of this substantially randomizing the observations if compared across different episodes. 

\begin{figure}[h]
\centering
\includegraphics[width=\linewidth]{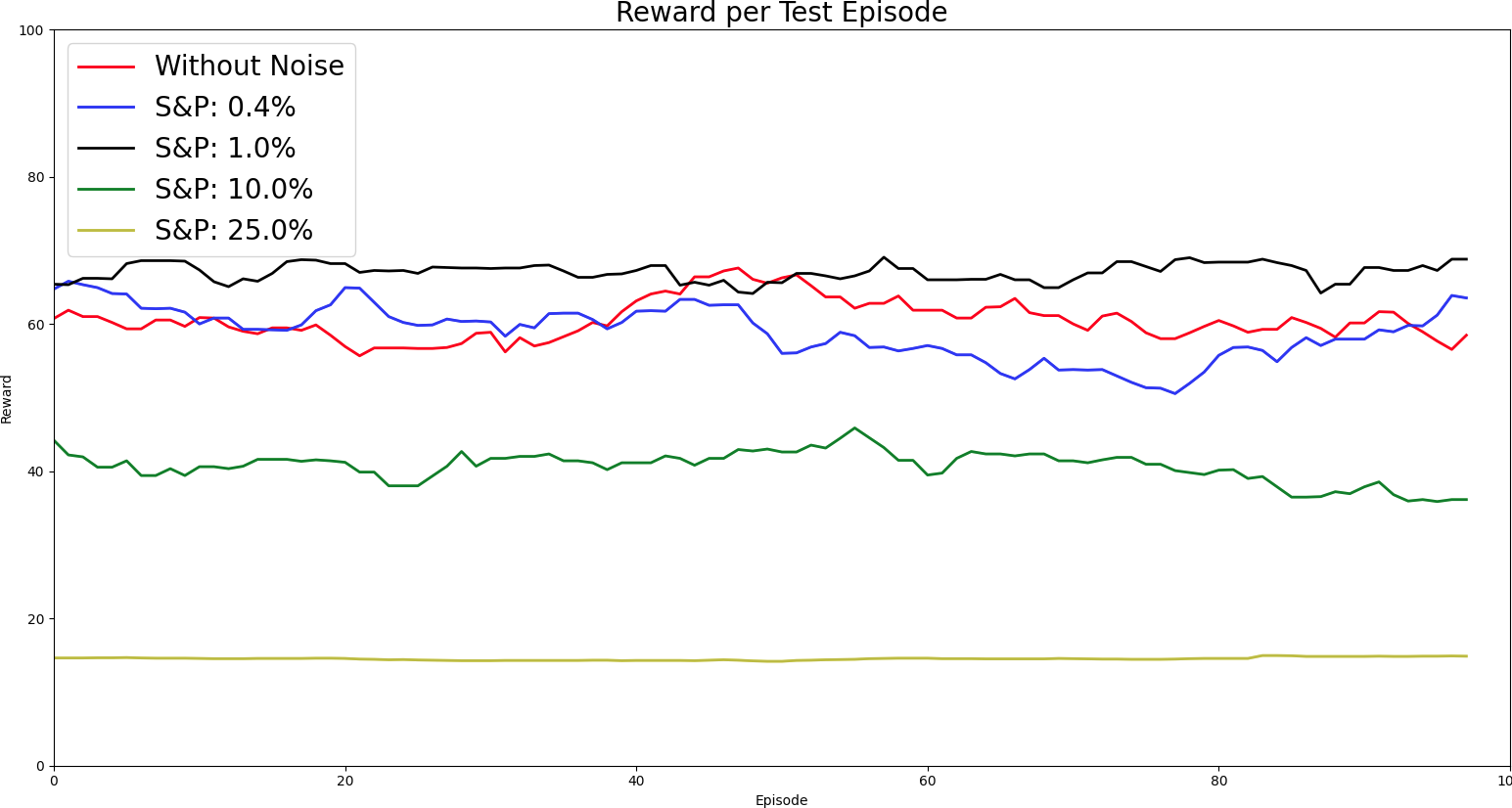}
\caption{Comparison of the moving average rewards performed in $100$ test episodes.}
\label{fig:reward_comparition}
\end{figure}
\vspace{-2mm}

\section{Conclusion}

In this paper, we have proposed a DRL-based approach for training an agent to play a visually complex modern video game. The method only uses the image as input and the results obtained in the trained scenario shown that the agent was capable to perform in the proposed scenario successfully and learn to avoid the obstacles to reach the desired target.

More specifically, we can examine that with a simple low-dimensional observation filter structure, the DRL algorithm obtained a good performance. It managed to avoid the obstacles and execute a successful route in the environment. We can also conclude that DRL approaches are suitable for the development of systems that require discrete control for artificial agents, in a visually complex environment using the image as observation. The robust learning results obtained can also be attributed to our rewarding scheme being simple and precise, such as those proposed.

In future works, we aim to expand the study with more DRL algorithms, such as Deep Deterministic Policy Gradient (DDPG) and Twin Delayed Deep Deterministic (TD3). We also intend to test our approach in more modern video games.

\section*{Acknowledgement}

The authors would like to thank the VersusAI team.

\bibliographystyle{IEEEtran}
\bibliography{IEEEabrv,IEEEexample}

\begin{thebibliography}{10}
\providecommand{\url}[1]{#1}
\csname url@samestyle\endcsname
\providecommand{\newblock}{\relax}
\providecommand{\bibinfo}[2]{#2}
\providecommand{\BIBentrySTDinterwordspacing}{\spaceskip=0pt\relax}
\providecommand{\BIBentryALTinterwordstretchfactor}{4}
\providecommand{\BIBentryALTinterwordspacing}{\spaceskip=\fontdimen2\font plus
\BIBentryALTinterwordstretchfactor\fontdimen3\font minus
  \fontdimen4\font\relax}
\providecommand{\BIBforeignlanguage}[2]{{%
\expandafter\ifx\csname l@#1\endcsname\relax
\typeout{** WARNING: IEEEtran.bst: No hyphenation pattern has been}%
\typeout{** loaded for the language `#1'. Using the pattern for}%
\typeout{** the default language instead.}%
\else
\language=\csname l@#1\endcsname
\fi
#2}}
\providecommand{\BIBdecl}{\relax}
\BIBdecl

\bibitem{skinner2019icccs}
G.~Skinner and T.~Walmsley, ``Artificial intelligence and deep learning in
  video games a brief review,'' in \emph{2019 IEEE 4th International Conference
  on Computer and Communication Systems (ICCCS)}, 2019, pp. 404--408.

\bibitem{yannakakis2018artificial}
\BIBentryALTinterwordspacing
G.~N. Yannakakis and J.~Togelius, \emph{Artificial intelligence and
  games}.\hskip 1em plus 0.5em minus 0.4em\relax Springer, 2018, vol.~2.
  [Online]. Available:
  \url{https://link.springer.com/content/pdf/10.1007/978-3-319-63519-4.pdf}
\BIBentrySTDinterwordspacing

\bibitem{mnih2015human}
\BIBentryALTinterwordspacing
V.~Mnih, K.~Kavukcuoglu, D.~Silver, A.~A. Rusu, J.~Veness, M.~G. Bellemare,
  A.~Graves, M.~Riedmiller, A.~K. Fidjeland, G.~Ostrovski \emph{et~al.},
  ``Human-level control through deep reinforcement learning,'' \emph{nature},
  vol. 518, no. 7540, pp. 529--533, 2015. [Online]. Available:
  \url{https://www.nature.com/articles/nature14236?wm=book_wap_0005}
\BIBentrySTDinterwordspacing

\bibitem{bellemare2013arcade}
\BIBentryALTinterwordspacing
M.~G. Bellemare, Y.~Naddaf, J.~Veness, and M.~Bowling, ``The arcade learning
  environment: An evaluation platform for general agents,'' \emph{Journal of
  Artificial Intelligence Research}, vol.~47, pp. 253--279, 2013. [Online].
  Available: \url{https://www.jair.org/index.php/jair/article/view/10819}
\BIBentrySTDinterwordspacing

\bibitem{arulkumaran2017deep}
K.~Arulkumaran, M.~P. Deisenroth, M.~Brundage, and A.~A. Bharath, ``Deep
  reinforcement learning: A brief survey,'' \emph{IEEE Signal Processing
  Magazine}, vol.~34, no.~6, pp. 26--38, 2017.

\bibitem{laskin2020reinforcement}
M.~Laskin, K.~Lee, A.~Stooke, L.~Pinto, P.~Abbeel, and A.~Srinivas,
  ``Reinforcement learning with augmented data,'' \emph{arXiv preprint
  arXiv:2004.14990}, 2020.

\bibitem{torrado2018deep}
R.~R. Torrado, P.~Bontrager, J.~Togelius, J.~Liu, and D.~Perez-Liebana, ``Deep
  reinforcement learning for general video game ai,'' in \emph{2018 IEEE
  Conference on Computational Intelligence and Games (CIG)}.\hskip 1em plus
  0.5em minus 0.4em\relax IEEE, 2018, pp. 1--8.

\bibitem{ha2018world}
\BIBentryALTinterwordspacing
D.~Ha and J.~Schmidhuber, ``World models,'' \emph{Conference on Neural
  Information Processing Systems}, 2018. [Online]. Available:
  \url{https://arxiv.org/abs/1803.10122}
\BIBentrySTDinterwordspacing

\bibitem{ha2018recurrent}
\BIBentryALTinterwordspacing
------, ``Recurrent world models facilitate policy evolution,''
  \emph{Conference on Neural Information Processing Systems}, 2018. [Online].
  Available: \url{https://arxiv.org/abs/1809.01999}
\BIBentrySTDinterwordspacing

\bibitem{shao2019survey}
\BIBentryALTinterwordspacing
K.~Shao, Z.~Tang, Y.~Zhu, N.~Li, and D.~Zhao, ``A survey of deep reinforcement
  learning in video games,'' 2019. [Online]. Available:
  \url{https://arxiv.org/abs/1912.10944}
\BIBentrySTDinterwordspacing

\bibitem{vinyals2017starcraft}
\BIBentryALTinterwordspacing
O.~Vinyals, T.~Ewalds, S.~Bartunov, P.~Georgiev, A.~S. Vezhnevets, M.~Yeo,
  A.~Makhzani, H.~K{\"u}ttler, J.~Agapiou, J.~Schrittwieser \emph{et~al.},
  ``Starcraft ii: A new challenge for reinforcement learning,'' \emph{arXiv
  preprint arXiv:1708.04782}, 2017. [Online]. Available:
  \url{https://arxiv.org/abs/1708.04782}
\BIBentrySTDinterwordspacing

\bibitem{berner2019dota}
\BIBentryALTinterwordspacing
C.~Berner, G.~Brockman, B.~Chan, V.~Cheung, P.~Debiak, C.~Dennison, D.~Farhi,
  Q.~Fischer, S.~Hashme, C.~Hesse \emph{et~al.}, ``Dota 2 with large scale deep
  reinforcement learning,'' \emph{arXiv preprint arXiv:1912.06680}, 2019.
  [Online]. Available: \url{https://arxiv.org/abs/1912.06680}
\BIBentrySTDinterwordspacing

\bibitem{tessler2017deep}
C.~Tessler, S.~Givony, T.~Zahavy, D.~Mankowitz, and S.~Mannor, ``A deep
  hierarchical approach to lifelong learning in minecraft,'' in
  \emph{Proceedings of the AAAI Conference on Artificial Intelligence},
  vol.~31, no.~1, 2017.

\bibitem{lin2019using}
C.-J. Lin, J.-Y. Jhang, H.-Y. Lin, C.-L. Lee, and K.-Y. Young, ``Using a
  reinforcement q-learning-based deep neural network for playing video games,''
  \emph{Electronics}, vol.~8, no.~10, p. 1128, 2019.

\bibitem{tan2019modeling}
R.~Tan, J.~Zhou, H.~Du, S.~Shang, and L.~Dai, ``An modeling processing method
  for video games based on deep reinforcement learning,'' in \emph{2019 IEEE
  8th joint international information technology and artificial intelligence
  conference (ITAIC)}.\hskip 1em plus 0.5em minus 0.4em\relax IEEE, 2019, pp.
  939--942.

\bibitem{kaiser2019model}
L.~Kaiser, M.~Babaeizadeh, P.~Milos, B.~Osinski, R.~H. Campbell, K.~Czechowski,
  D.~Erhan, C.~Finn, P.~Kozakowski, S.~Levine \emph{et~al.}, ``Model-based
  reinforcement learning for atari,'' \emph{arXiv preprint arXiv:1903.00374},
  2019.

\bibitem{dechter1986learning}
R.~Dechter, ``Learning while searching in constraint-satisfaction problems,''
  \emph{Association for the Advancement of Artificial Intelligence}, 1986.

\bibitem{schmidhuber2015deep}
J.~Schmidhuber, ``Deep learning in neural networks: An overview,'' \emph{Neural
  networks}, vol.~61, pp. 85--117, 2015.

\bibitem{krizhevsky2012imagenet}
\BIBentryALTinterwordspacing
A.~Krizhevsky, I.~Sutskever, and G.~E. Hinton, ``Imagenet classification with
  deep convolutional neural networks,'' \emph{Advances in neural information
  processing systems}, vol.~25, pp. 1097--1105, 2012. [Online]. Available:
  \url{https://kr.nvidia.com/content/tesla/pdf/machine-learning/imagenet-classification-with-deep-convolutional-nn.pdf}
\BIBentrySTDinterwordspacing

\bibitem{lecun2015deep}
\BIBentryALTinterwordspacing
Y.~LeCun, Y.~Bengio, and G.~Hinton, ``Deep learning,'' \emph{nature}, vol. 521,
  no. 7553, pp. 436--444, 2015. [Online]. Available:
  \url{https://www.nature.com/articles/nature14539}
\BIBentrySTDinterwordspacing

\bibitem{sutton2018reinforcement}
R.~S. Sutton and A.~G. Barto, \emph{Reinforcement learning: An
  introduction}.\hskip 1em plus 0.5em minus 0.4em\relax MIT press, 2018.

\bibitem{mnih2013playing}
\BIBentryALTinterwordspacing
V.~Mnih, K.~Kavukcuoglu, D.~Silver, A.~Graves, I.~Antonoglou, D.~Wierstra, and
  M.~A. Riedmiller, ``Playing atari with deep reinforcement learning,''
  \emph{NIPS Deep Learning Workshop}, vol. abs/1312.5602, 2013. [Online].
  Available: \url{http://arxiv.org/abs/1312.5602}
\BIBentrySTDinterwordspacing

\bibitem{lillicrap2015continuous}
\BIBentryALTinterwordspacing
T.~P. Lillicrap, J.~J. Hunt, A.~Pritzel, N.~Heess, T.~Erez, Y.~Tassa,
  D.~Silver, and D.~Wierstra, ``Continuous control with deep reinforcement
  learning,'' in \emph{4th Int. Conf. on Learning Representations, {ICLR}},
  Y.~Bengio and Y.~LeCun, Eds., 2016. [Online]. Available:
  \url{http://arxiv.org/abs/1509.02971}
\BIBentrySTDinterwordspacing

\bibitem{bar2005image}
L.~Bar, N.~Sochen, and N.~Kiryati, ``Image deblurring in the presence of
  salt-and-pepper noise,'' in \emph{International Conference on Scale-Space
  Theories in Computer Vision}.\hskip 1em plus 0.5em minus 0.4em\relax
  Springer, 2005, pp. 107--118.

\end{thebibliography}

\end{document}